\documentclass[10pt,conference,a4paper]{IEEEtran}
\IEEEoverridecommandlockouts

\usepackage{times}
\usepackage{graphicx}
\usepackage{latexsym}
\usepackage{amsmath,amssymb,amsfonts}
\usepackage{multirow}
\usepackage{float}
\usepackage{textcomp}
\usepackage[misc]{ifsym}
\usepackage{cite}
\usepackage{color}
\usepackage{tablefootnote}
\usepackage{threeparttable}
\usepackage{changes}

\newcommand{\fin}{\color{black}}

\usepackage{ifpdf}
\usepackage{algorithmic}

\usepackage{array}

\ifCLASSOPTIONcompsoc
  \usepackage[caption=false,font=normalsize,labelfont=sf,textfont=sf]{subfig}
\else
  \usepackage[caption=false,font=footnotesize]{subfig}
\fi
\usepackage{url}

\begin{document}
%
\title{Three-Dimensional Lip Motion Network for Text-Independent Speaker Recognition}



%

\author{\IEEEauthorblockN{Jianrong Wang$^{1,2}$,
Tong Wu$^1$,
Shanyu Wang$^1$,
Mei Yu$^{1,2}$,
Qiang Fang$^3$
Ju Zhang$^1$ and
Li Liu$^{4^{\ast}}$\thanks{* Corresponding author}}

\IEEEauthorblockA{$^1$College of Intelligence and Computing, Tianjin University, China.
E-mail: {wjr, wutong18, sywang, yumei, juzhang}@tju.edu.cn}
\IEEEauthorblockA{$^2$Tianjin Key Laboratory of Cognitive Computing and Application, China}
\IEEEauthorblockA{$^3$Institute of Linguistics,
Chinese Academy of Social Sciences, China. E-mail: fangqiang@cass.org.cn}
\IEEEauthorblockA{$^4$Shenzhen Research Institute of Big Data, the chinese university of hong kong shenzhen, China. E-mail: liuli@cuhk.edu.cn}}



\maketitle

\begin{abstract}

Lip motion reflects behavior characteristics of speakers, and thus can be used as a new kind of biometrics in speaker recognition. In the literature, lots of works used two-dimensional (2D) lip images to recognize speaker in a text-dependent context. However, 2D lip easily suffers from various face orientations. To this end, in this work, we present a novel end-to-end 3D lip motion Network (3LMNet) by utilizing the sentence-level 3D lip motion (S3DLM) to recognize speakers in both the text-independent and text-dependent contexts. A new regional feedback module (RFM) is proposed to obtain attentions in different lip regions. Besides, prior knowledge of lip motion is investigated to complement RFM, where landmark-level and frame-level features are merged to form a better feature representation. Moreover, we present two methods, i.e., coordinate transformation and face posture correction to pre-process the LSD-AV dataset, which contains 68 speakers and 146 sentences per speaker. The evaluation results on this dataset demonstrate that our proposed 3LMNet is superior to the baseline models, i.e., LSTM, VGG-16 and ResNet-34, and outperforms the state-of-the-art using 2D lip image as well as the 3D face. The code of this work is released at  \emph{https://github.com/wutong18/Three-Dimensional-Lip-Motion-Network-for-Text-Independent-Speaker-Recognition}\fin.
\end{abstract}


%
\IEEEpeerreviewmaketitle

\section{Introduction}
With the development of automation technology, identity technology is widely used in various applications, such as financial transaction identity verification \cite{8398930, 7977654}, security access control \cite{8858704, 8716452, 8266457}, human-computer interaction \cite{8611591, 8329907}. At the beginning, identification mainly depends on the password authentication, then it has been developed to use biometric identification methods such as face recognition and fingerprint recognition. Articulatory features are also a biometric method that can be used for identification \cite{8893188, 8845454}, since speaking behaviors have personal habits and are difficult to imitate.



Lip motion is regarded as a behavioral feature, and is affected by the speaker's speaking habits and differs from person to person. It shows that human lip alone is able to provide sufficient information related to the identity of its owner\cite{1510631}. In recent years, speaker recognition methods using lip motion have begun to appear \cite{7533099, XinjunSpeaker, Yan2017, liu2018visual}. Compared with traditional biological features, lip motion can improve security by exploiting the dynamic information \cite{5203875, 6890423}.

Lip motion-based speaker recognition usually uses 2D images to extract lip motion features\cite{liu2017inner}. Aravabhumi et al. \cite{5598333} presented a system for extracting and analyzing lip motion features, which tackled the problem of open-set speaker identification. Meng et al. \cite{6019829} used time-sequential images to build constructed Hidden Markov Models (HMMs) of lip movements, and showed an effective performance. Shi et al.\cite{7533099} introduced a confident measure to discriminate three kinds of speakers' lip feature and got a reliable authentication result. However, if the 2D images are not photographed in a correct orientation., the speaker recognition will be difficult\cite{8451204}.

To provide more identified information than 2D data and address the variations caused by different speaker’s pose and position, researchers have proposed using 3D data with depth information\cite{6207656}. To the best of our knowledge, few work uses only 3D lips motion in speaker recognition, except that some works \cite{6207656, 7532912} use 3D faces. However, a 3D face contains a large number of landmarks (1347 points derived by Kinect V2), and some facial information may be redundant.

\begin{figure*}[!t]
	\centering
	\includegraphics[width=0.95\textwidth]{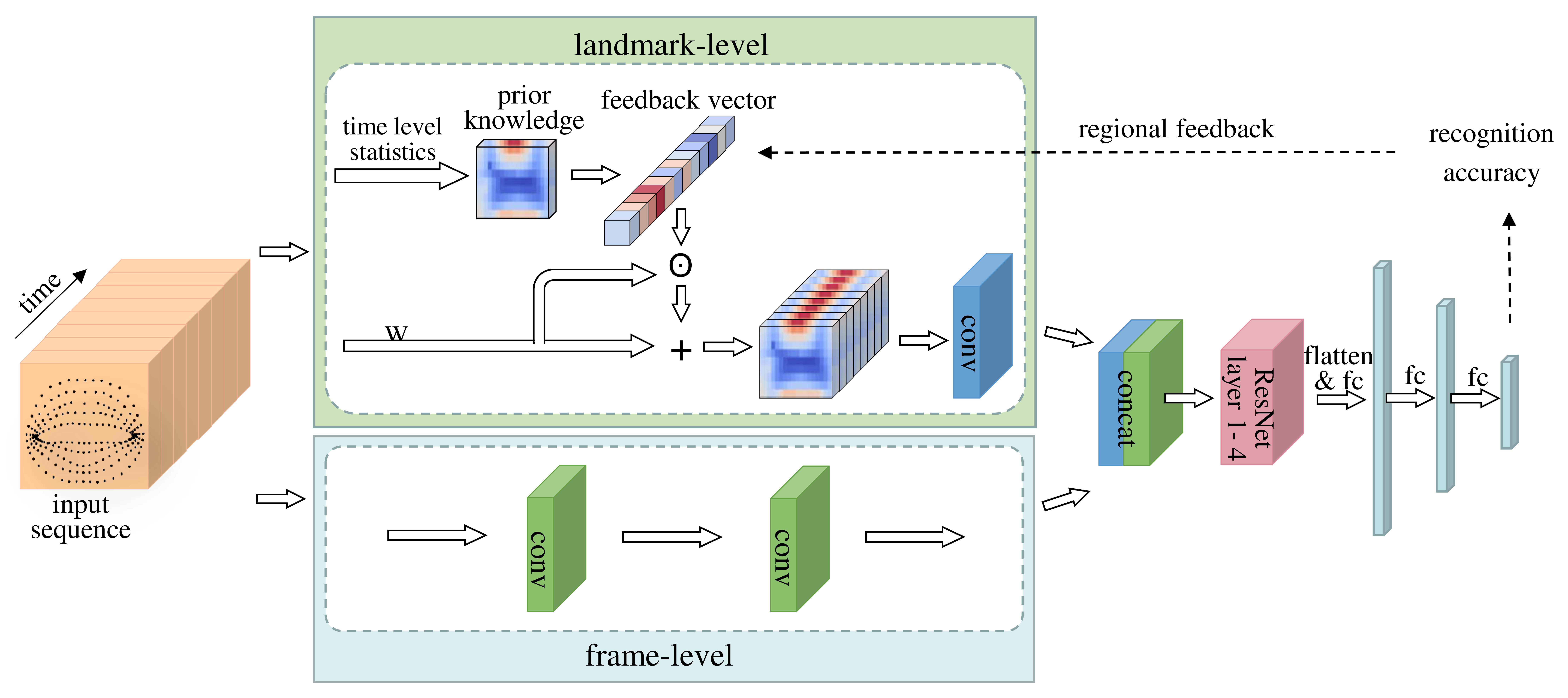}
	\caption{Overview of the proposed 3LMNet.}
	\label{figure3_overview}
\end{figure*}

To avoid using the whole face information, and to explore the potential of using lip information only in speaker recognition, in this paper, we present a novel 3D lip motion Network (3LMNet) by proposing a Regional Feedback Module (RFM) to screen out key identifying information in lip motion. Besides, a prior knowledge of lip motion is introduced in the RFM, which greatly improves the recognition performance. We formulate a new lip data representation, which is a sentence-level 3D lip motion (S3DLM) sequence, describing the lip shape by a number of 3D landmarks. Besides, data preprocessing procedures are investigated to correct the S3DLM sequences through coordinate transformation and posture correction procedures. Our methods are evaluated on a Mandarin Dataset \cite{WangCorpus}, which is a large-scale depth-based multimodal audio-visual corpus, including 3D lip points of 68 speakers. For convenience, we name it LSD-AV dataset in this work. The evaluation results on this dataset demonstrate that our proposed 3LMNet outperforms three baseline models and the state-of-the-art (SOTA) \cite{8451204} by a margin of 1.99\%.
The overview of our proposed 3LMNet is shown in Fig. \ref{figure3_overview}.

The contributions of this work can be summarized as follows:

\begin{enumerate}
\item To the best of our knowledge, this is the first work that uses S3DLM sequences in the speaker recognition task.
\item We present a novel 3LMNet based on S3DLM sequences by proposing the RFM and incorporating the prior knowledge of speaker's lip motion.
\item Evaluation results on the LSD-AV dataset show that our proposed 3LMNet is superior to the baseline models, i.e., LSTM, VGG-16 and ResNet-34, and outperforms the SOTA performance in \cite{8451204}.

\end{enumerate}

\section{Related Work}
Speaker recognition using lip motion has been studied for a long time. In previous studies, lip motion was mainly based on 2D images. Roland et al.\cite{Auckenthaler2015Lip} analyzed the movement contours of the upper and lower lip during speaking, and evaluated the visual lip features to recognize speakers.  Ma et al.\cite{XinjunSpeaker} proposed to use a fixed password (i.e., text-dependent) lip motion images to verify the speakers. Liao et al.\cite{8451204} proposed 3D convolutional neural network (CNN), which alleviates the influence caused by variations of speaker's poses and positions. Aravabhumi et al. \cite{5598333} measured the distance and angle from the lip centroid to all boundary lip points in lip images to identify the speaker.

With the popularity of depth cameras, 3D data has attracted people's attention since it can alleviate variations in illumination and head pose problems in 2D-based recognition. In literature, there are some works dedicated to the person recognition using 3D face. For example, Wu et al.\cite{7867242} used a face point cloud to recognize human faces, and Jhuang et al.\cite{7899838} adopted deep belief networks to train the 3D face point cloud, and achieved good results in person verification. However, these works used face information that is not related to speech. In addition, as far as we know, few works used 3D lip motion for speaker recognition.

On the other hand, we introduce two types of speaker recognition, i.e., text-dependent and text-independent. Text-dependent recognition specifies text content during training and test stages. Under such constraints, better results can be achieved, but users need to cooperate actively. Liu et al. \cite{XinLearning} proposed the concept of lip motion password and used HMMs to identify text-dependent speakers. Text-independent recognition does not specify text content, but feature extraction is more difficult. This recognition does not require users' cooperation and thus has a broader range of applications. Tatsuya et al.\cite{6603515} used lip movement of the same word to identify the speakers, and further found that different words also seem to have personal characteristics, but they are far inferior to the same word. Liao et al.\cite{4607501} proposed a lip motion-based speaker verification scheme using random passwords.

\section{Method}

\subsection{S3DLM Sequence}

A set of 3D coordinates can describe a sequence of lip motion.  In order to take advantage of the CNN model's excellent ability to extract high-level information, we convert the spatio-temporal data into a data form that can be fed to a 2D CNN structure. Besides, 2D convolution is used to process spatio-temporal information to avoid a large number of parameters using 3D convolutions.

\begin{figure}[!t]
\centering
\includegraphics[width=0.83\linewidth]{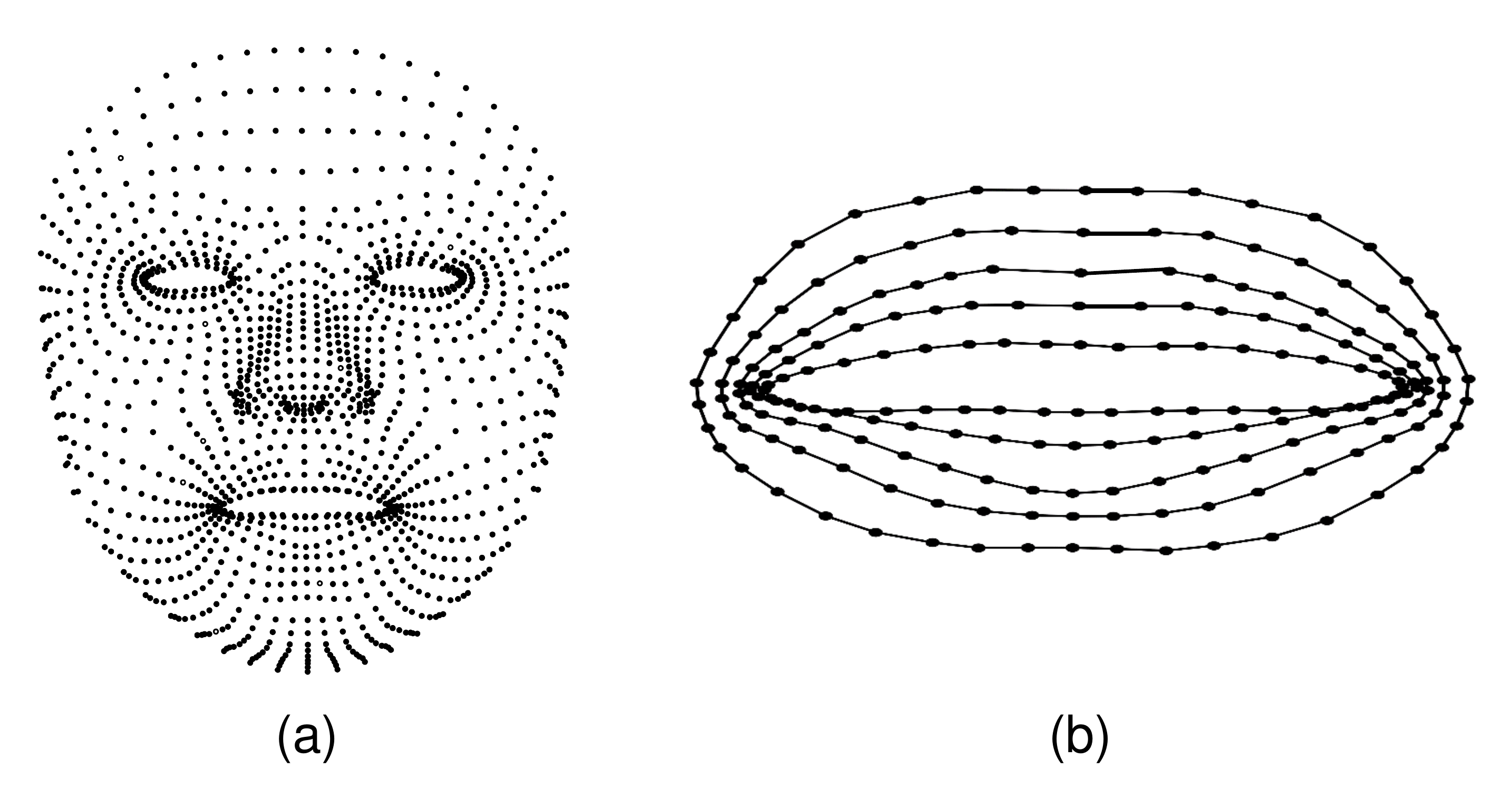}
\caption{(a) 3D facial point cloud that contains 1347 landmarks to describe the whole face, (b) 200 lip landmarks shown in 10 lip rows.}
\label{figure2_face}
\end{figure}

We selected 200 points of the lip from the 1347 facial landmarks of LSD-AV dataset, with 100 landmarks on the lower lip and upper lip, as shown in Fig. \ref{figure2_face}. For convenience, we define feature landmarks of the upper and lower lips as 5 lip rows, 20 landmarks per lip row. Since the length of sentences in the corpus is different, and the speaking speed of speakers is variable, we sort the lip data of each sentence by time, and then select 28 frames of 3D lip points in each sentence to represent the movements of lips.

To represent the spatial position of the lip and capture the dynamic temporal information, we specify the S3DLM sequences as the minimum feature units of network. More precisely, each sequence is represented by a lip point vector, i.e., $S=(P_1, P_2, \ldots, P_t, \ldots, P_{28})$, where $P_t$ indicates lip state at $t$ frame. $P_t=(P_t^1, P_t^2, \ldots, P_t^k, \ldots, P_t^{200})$ with $P_t^k=(x_t^k, y_t^k, z_t^k)$, and $k$ is the serial number of lip point. Specifically, in this work, one sequence can be represented by a $28\times200\times3$ tensor.

The S3DLM sequences not only show the static spatial position of lip landmarks, but also represent the dynamic lip motion in adjacent frames. It is beneficial for the network to directly extract the static features of the lip and compare the personalized dynamic features of the lip motion.

\subsection{The Relationship between Text and Lip Motion Characteristics}
The lip motion of different speakers is driven by the same lip muscles, and the lip motion is similar \cite{King2001A}. However, due to differences in speaking habits and lip muscle physiological characteristics among people, the movements in the lip area are different in direction, amplitude, and frequency\cite{5438780}. On this basis, lip movements are mainly related to the text contents, speakers' lip muscle physiological characteristics and their speaking habits. To explore the relationship between text content and lip motion, we have $(1)$ as follows.

\begin{equation}
P_{ij}=u_{j}+l_i+\varepsilon_{ij},
\label{equ0}
\end{equation}
where $u_j$ is the text-dependent lip motion vector, which is mainly determined by the content of the speech, and $j$ is the text content. $l_i$ is the personalized lip motion vector related with speaker habits and facial muscle differences, and $i$ is the speaker index. According to the central limit theorem, the mean value of the noise $\varepsilon_{ij}$ follows a normal distribution as follows
\begin{equation}
\frac{1}{n}\sum_{j=0}^{n}\varepsilon_{ij}=0, \quad \frac{1}{mn}\sum_{i=0}^{m}\sum_{j=0}^{n}\varepsilon_{ij}=0.
\label{equ4}
\end{equation}

To analyze the relationship between the individual differences for speaker lip motion and the text,
 we are interested in the variance of the speaker's lip motion $f$ as shown in $(3)$, and aim to prove that $f$ does not dependent on text content.

\begin{equation}
f=(\overline{P_i}-\overline{P})^2.
\label{equ1}
\end{equation}

The standard speaker's lip motion $\overline{P}$ in $(4)$ is a constant, and it is a common lip motion that is not relevant to the speakers' ID nor text contents.

\begin{equation}
\overline{P}=\frac{1}{m}\sum_{i=0}^{m}\overline{P_i}
            =\overline{u}+\overline{l}.
\label{equ3}
\end{equation}

By averaging all the texts in the dataset, we obtain $\overline{P_i}$ as follows
\begin{equation}
\overline{P_i}=\frac{1}{n}\sum_{j=0}^{n}(u_j+l_i+\varepsilon_{ij})
              =\overline{u}+l_i.
\label{equ2}
\end{equation}

We can see that $\overline{P_i}$ only depends on speaker's personal characteristics $l_i$.
According to $(1) - (5)$,
\begin{equation}
f=(l_i-\frac{\sum_{i=0}^{m}l_i}{m})^2.
\label{equ6}
\end{equation}
From (\ref{equ6}), we can conclude that the variance of the speaker's lip motion $f$ in (\ref{equ6})  only depends on the speaker, and thus is text-independent. Therefore, we theoretically prove that  speakers can be recognized based on text-independent lip motion.

\subsection{Regional Feedback Module}
To deal with the complexity of lip movement, we propose the RFM (shown in Fig. \ref{figure3_overview}) by treating lip feature landmarks as lip regions, and analyze the contribution of each region in the speaker recognition. To focus on informative regions and elimimate useless information, key landmarks are selected based on iterated feedback attention weights.
\begin{figure}[!t]
\centering
\includegraphics[width=2.2in]{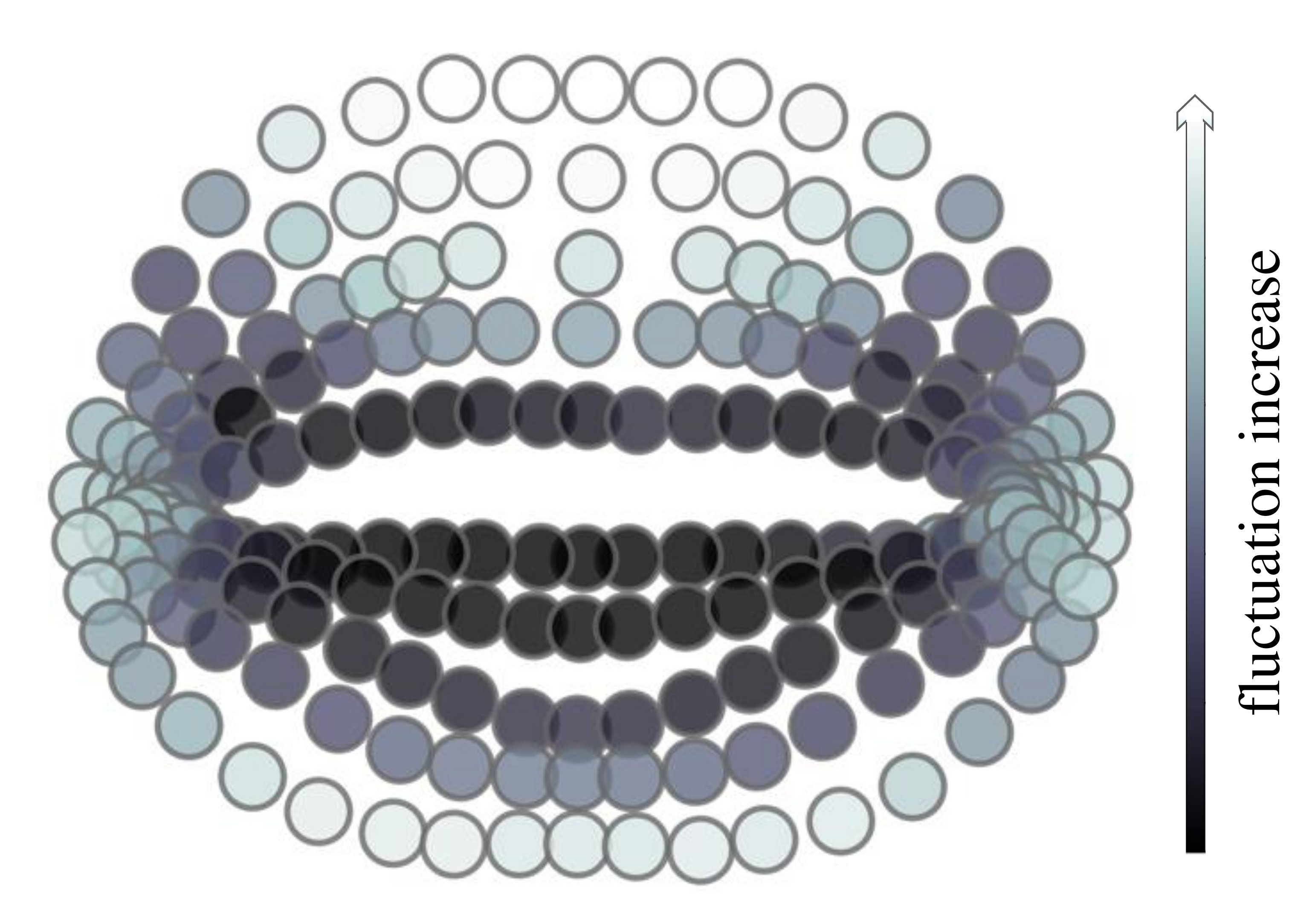}
\caption{The fluctuation of lip points. It is noted that the whiter of the points, the greater the fluctuation of the points are, and the closer to black, the smaller the points are.}
\label{figure4_flu}
\end{figure}

To further improve the efficiency and accuracy of data representation, we calculate prior knowledge of the lip motion based on 120 training sentences of the dataset. In detail, we first measure the deviation of lip landmarks from the reference (i.e., the average position of each lip landmark) as the prior knowledge of lip motion, which represents the fluctuation of lip points during speaking. Then apply the sigmoid function to map the deviation to the range (0, 1). We define the S3DLM sequence as $W$, where $W\in R^{m\times l}$, $m$ is the number of frames, and $l$ is the number of lip feature points. The value of $p$ can be accomplished via a logistic regression function:
\begin{equation}
p=sigmoid ( \frac{\alpha}{m} \sum_{i=0}^{m-1}\delta(W_i)+b),
\label{equ8}
\end{equation}
where $W_i$ is the lip points vector at $i$ frame, and $\delta$ is the variance function for each variable of the $W_i$. $\alpha$ and $b$ are the coefficients and offsets of the linear transformation, respectively.

We use the prior knowledge of lip movements as an initial value of the feedback vector to emphasize the movement of key lip points, which helps distinguish the contribution of lip points to the speaker recognition. It is defined as follows
\begin{equation}
\theta^{(p)}_0=p,
\label{equ9}
\end{equation}
where $\theta^{(p)}_0$ is the initial value of the feedback vector, representing the fluctuation of lip points. In the comparative experiment, we obtain the opposite prior knowledge using $1$ minus the prior knowledge. Fig. \ref{figure4_flu} shows the fluctuation of the 200 lip landmarks. We can see that fluctuation outside of the upper lip is the most obvious, followed by the outside of lower lip and the corners of mouth. The area with the least fluctuation is the inner lip.

The proposed RMF can be formulated as follows
\begin{equation}
\theta^{(p)}_{t+1}=\varphi [cov(W \odot  \theta^{(p)}_{t}+W),X_{frame}],
\label{equ7}
\end{equation}
where $\theta_{t}\in R^{l\times 1}$ is the feedback vector at epoch $t$, representing the contribution of lip regions to speaker recognition.
$\varphi$ is the feedback function of the recognition accuracy to the feedback vector, which includes convolution, flattening, and fully connected operations. $\odot$ represents the Hadamard product of each row of the matrix and a vector, which is implemented by multiplying each row of the matrix and the corresponding position element of the vector. $\theta^{(p)}_t$ and the original input $W$ are first subjected to Hadamard product, and then added to the original input $W$. After a convolution process, the landmark-level features are concatenated with the frame-level feature $X_{frame}$ and then sent to the network.

According to the recognition loss, the error information is propagated back. The feedback vector is updated according to the back-propagation of gradient on the loss function, as shown in the following,
\begin{equation}
\theta^{(p)}_{t+1}:=\theta^{(p)}_t-\alpha \frac{\partial}{\partial\theta^{(p)}_t}L,
\label{equ8}
\end{equation}
where $L$ is the loss function. We can see that the updated feedback vector affects the next iteration. Finally, the feedback vector obtains the best personalized information, which improves the recognition result.

\subsection{3D Lip Motion Network}
In this section, we describe our entire model, the 3LMNet, as shown in Fig. \ref{figure3_overview}. This end-to-end network learns landmark-level features and frame-level features of S3DLM sequences.

In the landmark-level feature stream,  a convolutional network is used to learn the lip motion features from the S3DLM sequences. The lip motion is first counted by time, and then used as the prior knowledge. The RFM adjusts the feedback vector according to the lip motion and recognition performance. The S3DLM sequences generate new sequences under the feedback vector as the input of the convolutional layer. In landmark-level feature extraction, the convolutional layer extracts the overall feature from multiple lip points in multiple frames.

In the frame-level stream, lip point is the extraction unit. We first extract the position features of lip points from the S3DLM sequences, and then extract the dynamic variations of each lip point in the adjacent frames. To simplify the training network, we use the 2nd to 5th layers of the ResNet-34 to train merged features. The mapping of the merged features is implemented by 3 fully connected (fc) layers.


\section{Experiment}

\subsection{Dataset}
The LSD-AV dataset\footnote{ The data used in this work is public (\emph{https://zenodo.org/record/3903974\#.XvG5X0YzbCI}).} in \cite{WangCorpus} used Microsoft Kinect V2 infrared sensor to obtain depth information through the time-of-flight method. The acquisition system collected depth information at a rate of 30 frames per second. The Windows software development kit was used to track the face in real-time and collected a 3D face point cloud composed of 1347 points. The acquired point cloud 3D coordinates represented the horizontal offset, vertical offset, and spatial distance between the acquired point and the infrared camera, respectively. The data set was based on a Chinese corpus, which lasted 22.4 hours of recording time totally. 69 speakers participated in the recording, and each speaker uttered 146 identical sentences.

In the text-independent experiment, the first 120 sentences of the corpus are used as the training set, and the remaining 26 sentences are used as the test set. In the text-dependent experiments, we randomly select 120 training sentences and 26 test sentences for each speaker. We adopt recognition accuracy to evaluate our model, which is the ratio of correctly recognized test samples and the total test samples.

\subsection{Data Preprocessing}
\subsubsection{Coordinate transformation}
The origin of the face point cloud coordinates is the center of the infrared camera. To facilitate the processing of the lip coordinates, we establish the mapping between the two coordinate systems by a coordinate axes translation. In the new coordinate system, the midpoint of the left and right mouth corners is the new coordinate origin. The line connecting to the mouth corners is the $X$ axis, the vertical direction above the corner of the mouth is the $Y$ axis, and the front of the face is the $Z$ axis.
\begin{figure}[!t]
	\centering
	\includegraphics[width=0.55\linewidth]{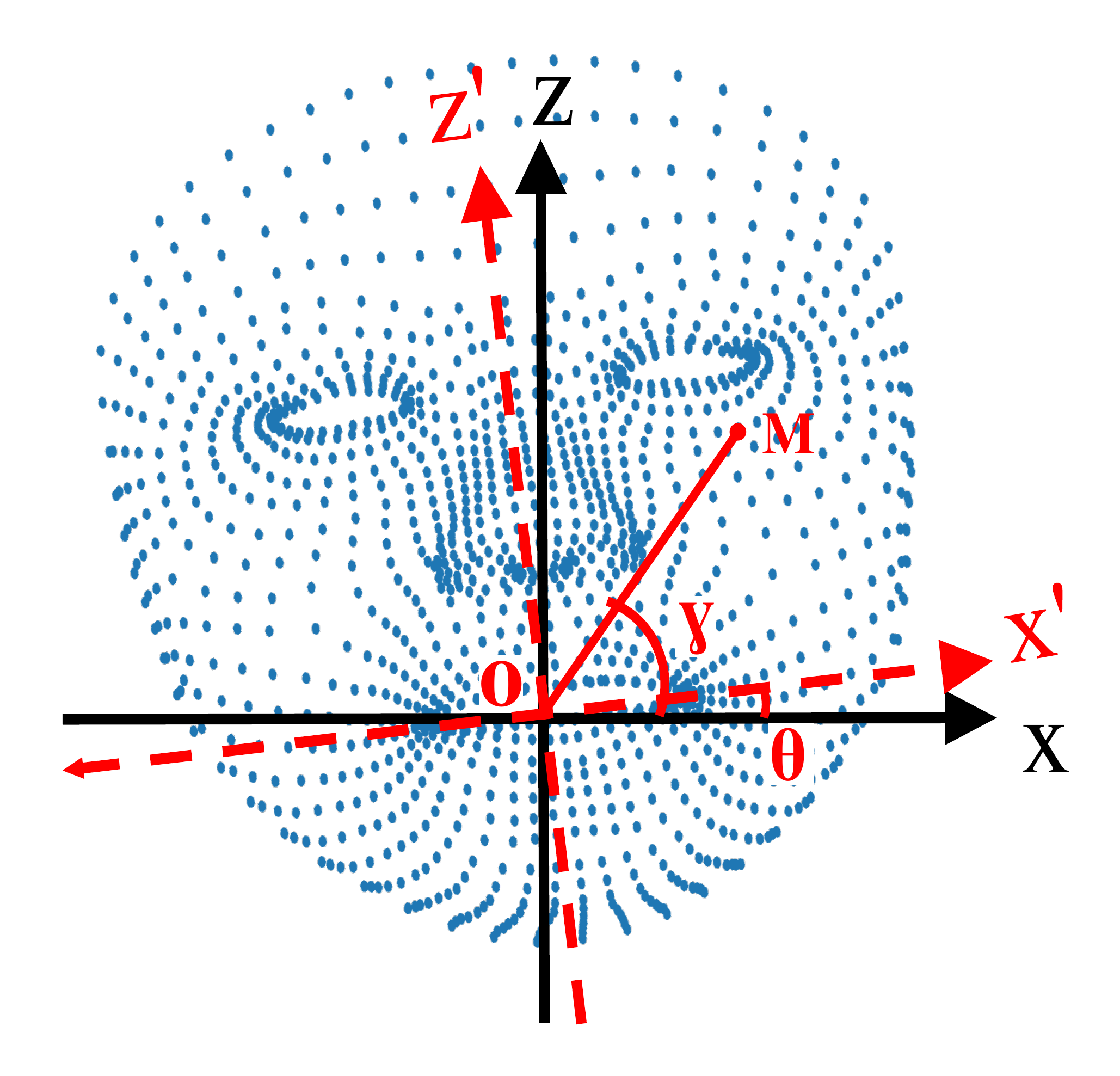}
	\caption{Feature point posture correction.}
	\label{figure4_coordinate}
\end{figure}
\subsubsection{Posture correction}
Since the speakers' head may rotate during the recording, we perform posture correction on the 3D point cloud, including  left/right rotation correction of the face, head tilt correction, and head up/down correction. Taking the rotation of a human face left and right as an example, as shown in Fig. \ref{figure4_coordinate}, the $XZ$ plane is the original coordinate system, the $X'Z'$ plane is the corrected coordinate system, and the origin O is the midpoint of mouth corners. The face point cloud rotates in the $ZX$ plane, while the $Y$ axis coordinate value is unchanged. The angle $\gamma$ between the point $M$ and the $X$ axis can be calculated by the coordinates of the original coordinate system:
\begin{equation}
\gamma=\frac{z_r-z_l}{x_r-x_l} ,
\label{equ9}
\end{equation}
where the coordinates of the corners of mouth in the original coordinate system are $(x_l, z_l)$ and $(x_r, z_r)$, respectively. The distance of two corners of mouth can be expressed as:

\begin{equation}
d^2=(x_r-x_l)^2+(z_r-z_l)^2.
\label{equ10}
\end{equation}

The coordinate of point $M$ in the original coordinate system  is $(x_m, z_m)$, which can be calculated by the distance of two corners of mouth and the angle between the point and the original coordinate axis,
\begin{equation}
x_m=d\cos\gamma, \quad z_m=d\sin\gamma.
\label{equ9}
\end{equation}

The coordinate of $M$ in the new coordinate system is $(x_{m'}, z_{m'})$, and $x_{m'}=r\cos(\gamma-\theta)$, $z_{m'}=r\sin(\gamma-\theta)$, where $\theta$ is the angle between the $X'-$ axis and the $X-$axis. The point coordinate transformation $M'=TM$ is obtained, where $T$ is a transformation matrix:
\begin{equation}
T=
\left(
    \begin{array}{ccc}
    \cos\theta & 0 & \sin\theta \\
    0 & 1 & 0 \\
    -\sin\theta & 0 & \cos\theta\\
    \end{array}
\right).
\label{equ10}
\end{equation}

Similarly, the head tilt and its posture can be corrected in the $XY$ plane and the $YZ$ plane. A visualization of the corrected lip point cloud is shown in Fig. \ref{figure2_face}(b).
\subsection{Implementation Details}

\subsubsection{Network structure}
In the 3LMNet, the convolutional kernel in the landmark-level stream is set to 5$\times$5, with padding size equals 2. ReLU activation is used to generate a point feature map with 32 channels. The first convolutional kernel in the frame-level stream is set to 1$\times$1 with no padding. The second convolutional kernel is set to 3$\times$1, with padding equals (1, 0), where 3 indicates adjacent frames. These two streams are trained simultaneously, and then are merged along channels. The dimensions of three fc layers are 1024, 256 and 68.

\subsubsection{Training and testing}
Our model is implemented by Pytorch on NVIDIA Tesla V100 GPU. The whole network is trained end-to-end using Adam with a learning rate of 0.01. We set the batch size to 100, the maximum iteration step to 1500, and decay the learning rates by a factor of 0.3 every 200 steps. In the fine-tuning stage, the learning rate is set to be 0.0001, and the number of iteration steps is 1000. The pre-training parameters of ResNet-34 are used as the initial parameters in the network.

\subsection{Result and analysis}
We first perform a statistical experiment to discuss the relationship between text and lip motion. Then, we conduct an experiment of text-independent speaker recognition to show the superiority of S3DLM sequences by comparing  S3DLM sequence and 2D landmarks.
At last, we carry out an ablation study to verify the influence of RFM and prior knowledge by both the text-independent and text-dependent speaker recognition.

\subsubsection{Discussion of the relationship between text and lip motion}
In Section \uppercase\expandafter{\romannumeral3}-B, we pointed out that $\overline{P_i}$ is the text-independent lip motion vector, thus text-independent lip motions can be used to recognize the speakers. To further verify this argument, we conduct a statistical experiment by taking the average of the text and the speakers on LSD-AV dataset, which contains 68 speakers and 146 sentences.

One is that we take the standard deviation (std) of the lip motion variance for every 20 sentences of all speakers as a sample (only the first 140 sentences are used). We denote them $\rm std_t$. The other is based on speakers, taking the std of the lip motion variance on all sentences of every 10 speakers as a sample (the last sample only contains 8 speakers). We denote them $\rm std_s$.
We examine the difference between $D_t$ (i.e., the difference between each element of $\rm std_t$ and its mean value) and the std of the 7 samples related to texts, and the $D_s$ (i.e., the difference between each element of $\rm std_s$ and the mean value) and the std of the 7 samples related to speakers in TABLE \uppercase\expandafter{\romannumeral1}.

We can see that the fluctuation of $D_t$ is not significant compared with its std (0.0046), while the fluctuation of $D_s$ is evident compared with its std (0.0431). It can be seen that the std of the $D_s$ is about 10 times larger than $D_t$. We experimentally show that the lip motion variance is less text-independent and more speaker-dependent, confirming that text-independent lip motion maintains personal characteristics and can be used to recognize speakers.



\begin{table*}[!t]
	\begin{center}
		{\caption{Comparison between text-based lip motion and speaker-based lip motion in the LSD-AV dataset.}
			\label{table1}}
		\begin{tabular}{ccccccccc}
			\hline
			\rule{0pt}{12pt}
			 Text-based sample ($\rm std_t$) &1-20&21-40 & 41-60 & 61-80 & 81-100 &101-120 &121-140 & std \\
			\hline
			\\[-6pt]
			$D_t$& 0.0067  & -0.0048 & 0.0024 & 0.0038 & -0.0017 & 0.0060 & 0.0004 & \bf 0.0046\\
			\hline
			\rule{0pt}{12pt}
			Speaker-based sample ($\rm std_s$) & 1-10  & 11-20 & 21-30 & 31-40 & 41-50 & 51-60 & 61-68  & std \\
			\hline
			\\[-6pt]
			$D_s$ & -0.0385 & 0.0293 & -0.0096 & 0.0738 & 0.0187 & -0.0465 & -0.0272 & \bf 0.0431\\
			\hline
		\end{tabular}
	\end{center}
\end{table*}


\subsubsection{Performance of the S3DLM sequences}
We present a text-independent speaker recognition experiment using the 3SLM sequences and 2D landmark sequences based on three benchmark models, i.e., LSTM\cite{HochreiterLong}, VGG-16\cite{Simonyan2014Very}, and ResNet-34\cite{7780459} to demonstrate the benefits of S3DLM sequence. Experimental results are shown in TABLE \uppercase\expandafter{\romannumeral2}.

We can see that the performance of using S3DLM sequences is better than that using 2D landmark sequences in all three models. Specifically, in the ResNet-34, the recognition accuracy using S3DLM sequences is 5.03\% higher than that using 2D landmarks. We hypothesize that 3D data can provide more stable and useful identification information than 2D data, and it is robust to various posture by applying the pose normalization to 3D data.

Besides, it shows that ResNet-34 outperforms slightly better than VGG-16, but outperforms LSTM by a large margin (over 10\%). This may be because that apart from the dynamic features of the lips, the spatial features of the lips are also important for speaker recognition. However, the LSTM pays more attention to extract the sequential dynamic information of the lips, while ignoring the spatial lip information, causing lower performance.

\begin{table}[!t]
	\begin{center}
		{\caption{Accuracy comparison of superior S3DLM sequence and 2D landmarks in different networks.}
			\label{table1}}
		\begin{tabular}{lcc}
			\hline
			\rule{0pt}{12pt}
			Model & S3DLM  & 2D landmarks\\
			\hline
			\\[-6pt]
			LSTM & 82.46\% & 76.23\% \\
			VGG-16 &  91.00\% & 87.10\% \\
			ResNet-34 & $\boldsymbol{93.50\%}$  & 88.47\%  \\
			\hline
		\end{tabular}
	\end{center}
\end{table}

\begin{figure*}[!t]
	\centering
	\subfloat[] {\includegraphics[width=0.3\textwidth]{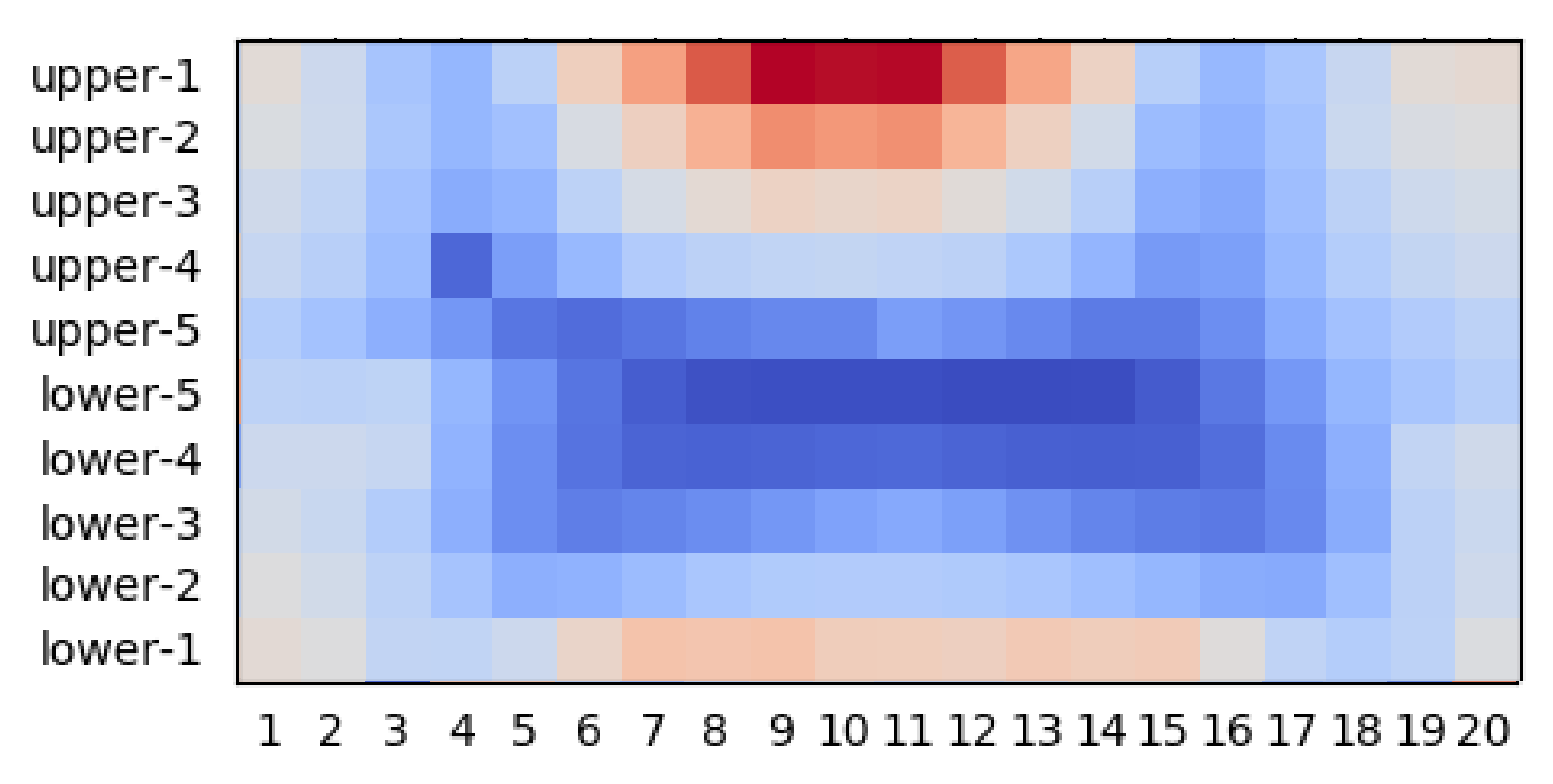}
		\label{1}}
	\hfil
	\subfloat[] {\includegraphics[width=0.3\textwidth]{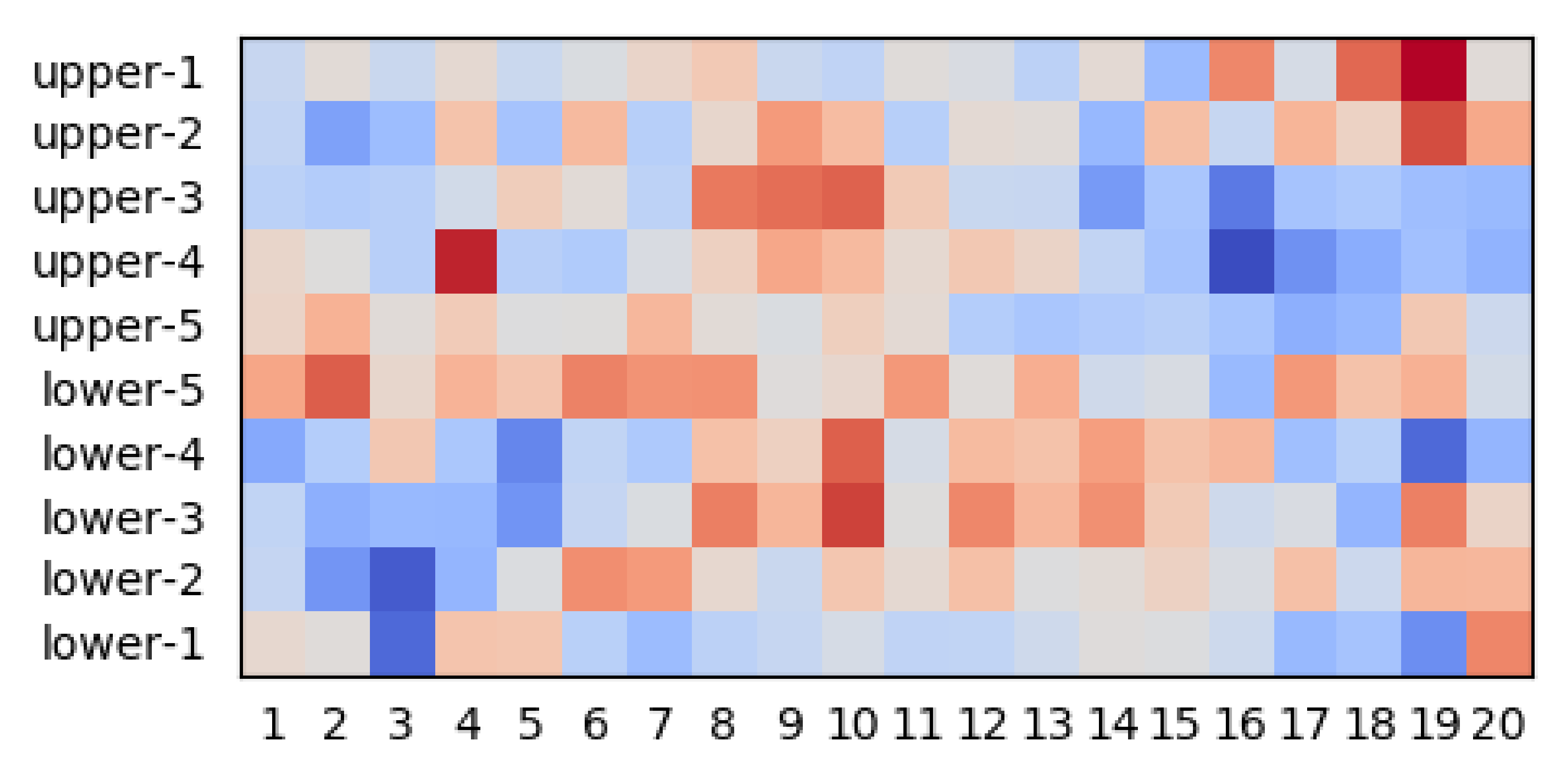}
		\label{2}}
	\hfil
	\subfloat[] {\includegraphics[width=0.3\textwidth]{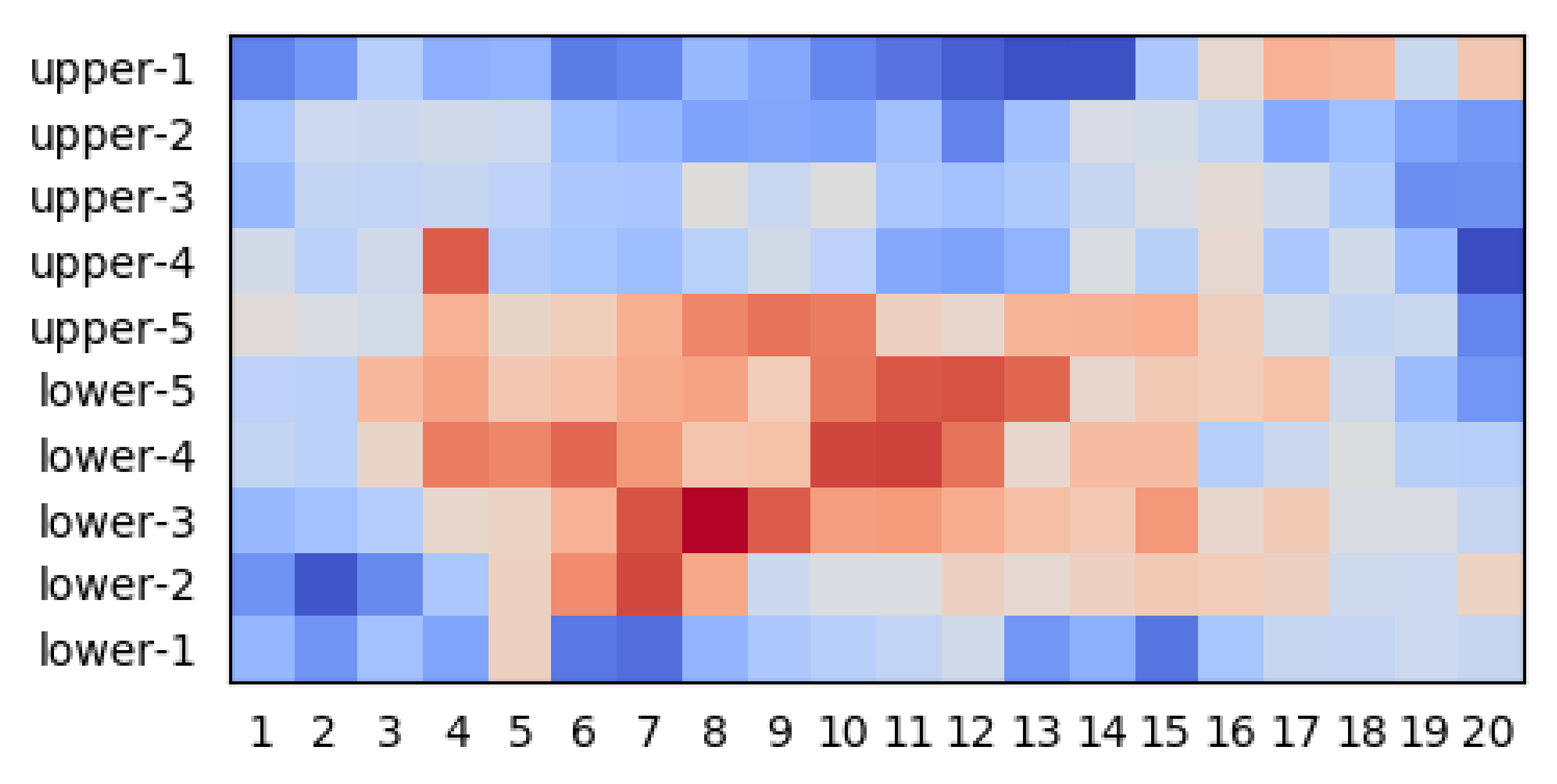}
		\label{3}}
	\caption{(a) Original lip points fluctuation; (b) Regional feedback visualization of lip points; (c) Regional feedback visualization with fluctuation prior knowledge of lip points. Note that in (a), the red lip points fluctuate significantly, while the blue points fluctuate less. In (b) and (c), the feedback module focuses on the lip points in the red area and the opposite in the blue area.}
	\label{figure8_prior}
\end{figure*}

\begin{table}[!t]
	\begin{center}
		{\caption{Ablation study of the proposed 3LMNet. The meaning of shorthand notations are listed in the Section \uppercase\expandafter{\romannumeral4}-D-3.}
			\label{table2}}
		\begin{threeparttable}
			\begin{tabular}{lcc}
				\hline
				\rule{0pt}{12pt}
				Model & Text-independent  & Text-dependent\\
				\hline
				\\[-6pt]
				Lai  et al.  \cite{LaiVisual}&---& 92.61\% \\
				Liao et al. \cite{8451204} &---& 97.11\% \\
				3LMNet-RFM-prior    & 93.91\%                    &  98.38\%  \\
				3LMNet+RFM-prior         & 94.94\%                    &  98.73\%  \\
				3LMNet+RFM+$\rm{prior_{opp}}$   & 91.94\%                    &  97.73\%  \\
				3LMNet+RFM+prior            & $\boldsymbol{95.22\%}$     &  $\boldsymbol{99.10\%}$  \\
				\hline
			\end{tabular}
		\end{threeparttable}
	\end{center}
\end{table}

\subsubsection{Performance of the 3LMNet}
Compared with the three benchmark models in TABLE \uppercase\expandafter{\romannumeral2}, we can see that our 3LMNet performs best. Compared with LSTM, the frame-level structure extracts the temporal features in three adjacent frames, which is similar with LSTM. However, our landmark-point level structure extracts the spatial features of the lip points, and personalized features, leading to a superior result to LSTM. Specifically, compared with the ResNet-34 model, the proposed RFM and prior knowledge make the 3LMNet to better identify the speaker from the key lip points of more personalized motion.

\paragraph{Ablation study}
To demonstrate the effectiveness of our proposed network and clearly explain how important roles of the key parts in the proposed 3LMNet, we compare the 3LMNet with and without RFM as well as the prior knowledge. Results on LSD-AV dataset are shown in TABLE \uppercase\expandafter{\romannumeral3}, where 3LMNet-RFM-prior is the baseline network without RFM and the prior knowledge of lip motion, 3LMNet+RFM-prior means using the RFM without prior, 3LMNet+RFM+$\rm{prior_{opp}}$ means using the RFM and the opposed prior, and 3LMNet+RFM+prior means using both the RFM and the prior knowledge. It shows that the proposed 3LMNet using both the RFM and the prior achieve the best performance, i.e., 95.22\% in text-independent context and 99.10\% in text-independent context.

In the verification of RFM, results based on 3LMNet+RFM-prior in text-independent and text-dependent speaker recognition are 1.03\% and 0.35\% higher than that using 3LMNet-RFM-prior, respectively. Due to the back-propagation, the feedback vector is updated according to the back propagation gradient of the loss function, as shown in (10). The RFM can distinguish the contribution of different lip regions from known lip motion to further capture the key lip regions of speaker recognition.

In the ablation experiment, recognition accuracies without the prior knowledge based on 3LMNet+RFM-prior in text-independent and text-dependent cases are reduced by 0.28\% and 0.37\% compared with 3LMNet+RFM+prior, respectively. The prior knowledge ablation experiment shows that the prior knowledge that emphasizes the personalized function has the best performance, while the performance of the prior knowledge that emphasizes the text is the worst. This is because prior knowledge is used as the initial value of the feedback vector to introduce the lip motion information. It is helpful to identify personalized lip points and help RFM adjust the contribution of speaker recognition.

To further analyze the feedback effect of RFM and the prior knowledge of lip motion, we visualize the feedback information, where Fig. \ref{figure8_prior}(a) shows the initial fluctuation statistics of lip landmarks, Fig. \ref{figure8_prior}(b) shows the visualization of the feedback vector with only RFM, and Fig. \ref{figure8_prior}(c) is the visualization of the feedback vector with both RFM and the prior. We can see that by using RFM, large weights are assigned to the regions with small fluctuations, which shows that the better the recognition effect, the lip regions with less variations have higher feedback weights. To verify this, we conducted a feedback comparison experiment (3LMNet+RFM+$\rm{prior_{opp}}$), where opposed prior is used. As shown in TABLE \uppercase\expandafter{\romannumeral3}, the comparison results are 3.28\% and 1.37\% lower in speaker recognition of text-independent and text-dependent compared with 3LMNet+RFM+prior, respectively. We conclude that strong fluctuations are mainly related to the text, which interferes with personal characteristics. In contrast, slight fluctuations are less affected by the text, and thus shows more obvious personality characteristics that are beneficial for speaker recognition.

\paragraph{Compare with the state-of-the-art}
To provide a comprehensive evaluation of the proposed S3DLM sequences and 3LMNet, we first compare our method with two SOTA works based on 2D images, i.e., Liao et al. \cite{8451204} and Lai et al. \cite{LaiVisual}. Since 3D lip motion is used in speaker recognition for the first time, we compare our method with these two text-dependent speaker recognition models using 2D lip image, which contains the entire lip image information, but it is sensitive to the pose rotations. The robustness of these two models are evaluated on different speaker-to-camera distances and various talking poses. The recognition accuracies that we listed are the average of all scenarios, which are 92.61\% and 97.11\%, respectively (see TABLE \uppercase\expandafter{\romannumeral3}), while our proposed 3LMNet with both RFM and prior achieves 99.10\%, outperforming these two works. More precisely, compared with 2D images,  the dimension and the needed amount of the lip key points are less by using S3DLM sequences. In \cite{8451204} and \cite{LaiVisual}, 125 frames of 300$\times$300 resolution lip images were used. Instead, we only use 200 lip point coordinates of 28 frames, while achieving better results.  


\section{Conclusion}

In this work, we propose a novel 3LMNet using 3D dynamic lip sequence in both text-independent and text-dependent speaker recognition. A novel RFM is designed to distinguish the contributions of different lip regions, and the prior knowledge derived from the known lip motion is investigated to further capture key lip regions. The experimental results on the LSD-AV dataset show the superior performance of our proposed 3LMNet compared with three baseline models, and achieve SOTA results in case of 2D lip motion speaker recognition as well as the 3D face verification. In future work, the Graph Convolutional Networks will be explored to capture better spatial lip information.

\section*{Acknowledgment}
This work was supported by the National Natural Science Foundation of China (No.61977049).

\bibliographystyle{IEEEtran}
\bibliography{IEEEabrv,ref}

\end{document}